\pdfoutput=1

\documentclass[11pt]{article}

\usepackage{acl}

\usepackage{times}
\usepackage{latexsym}

\usepackage[T1]{fontenc}

\usepackage{microtype}

\usepackage[disable]{todonotes}
\usepackage{amsmath}
\usepackage{amsfonts}
\usepackage{nert}
\usepackage{tikz-dependency}
\usepackage{adjustbox}
\usepackage{censor}

\makeatletter
\newcommand*\iftodonotes{\if@todonotes@disabled\expandafter\@secondoftwo\else\expandafter\@firstoftwo\fi}  
\makeatother

\title{Computational historical linguistics and language diversity in South Asia}

\author{Aryaman Arora \\
  Georgetown University \\
  \url{aa2190@georgetown.edu} \\\And
  Adam Farris \\
  San Mateo High School \\
  \url{adamfarris@gmail.com} \\\AND
  Samopriya Basu \\
  University of North Carolina -- Chapel Hill \\
  \url{sampr0b@live.unc.edu} \\\And
  Suresh Kolichala \\
  Microsoft \\
  \url{suresh.kolichala@gmail.com}}

\begin{document}
\maketitle
\begin{abstract}
South Asia is home to a plethora of languages, many of which severely lack access to new language technologies. This linguistic diversity also results in a research environment conducive to the study of comparative, contact, and historical linguistics---fields which necessitate the gathering of extensive data from many languages. We claim that data scatteredness (rather than scarcity) is the primary obstacle in the development of South Asian language technology, and suggest that the study of language history is uniquely aligned with surmounting this obstacle. We review recent developments in and at the intersection of South Asian NLP and historical--comparative linguistics, describing our and others' current efforts in this area. We also offer new strategies towards breaking the data barrier.
\end{abstract}

\section{Introduction}

South Asia\footnote{Roughly the Indian Subcontinent, or the geographic and cultural region enclosed by the Himalayas, the Indian Ocean, and the Hindu Kush.} is home to one-quarter of the world's population and boasts immense linguistic diversity \citep{saxena2008lesser,bashir}. With members of at least four top-level major linguistic families\footnote{Indo-European (Indo-Aryan, Iranic, Nuristani), Dravidian, Austroasiatic (Munda, Khasian), Sino-Tibetan (several branches).} and several putative linguistic isolates, this region is a fascinating arena for linguistic research. The languages of South Asia, moreover, have a long recorded history, and have undergone complex change through genetic descent, sociolinguistic interactions, and contact influence.

Nevertheless, South Asian languages for the most part remain severely underdocumented \citep{van2015endangered}, and several languages with even official administrative status (e.g.~Sindhi) are low-resource (if not data-scarce) for the purposes of all natural language processing tasks \citep{joshi-etal-2020-state}. This data scatteredness persists despite long native traditions of linguistic description, continued language vitality with active use on the internet, and vast numbers of speakers \citep{rahman2008language,groff2017language}.

\begin{figure}[t]
    \centering
    \includegraphics[width=\columnwidth]{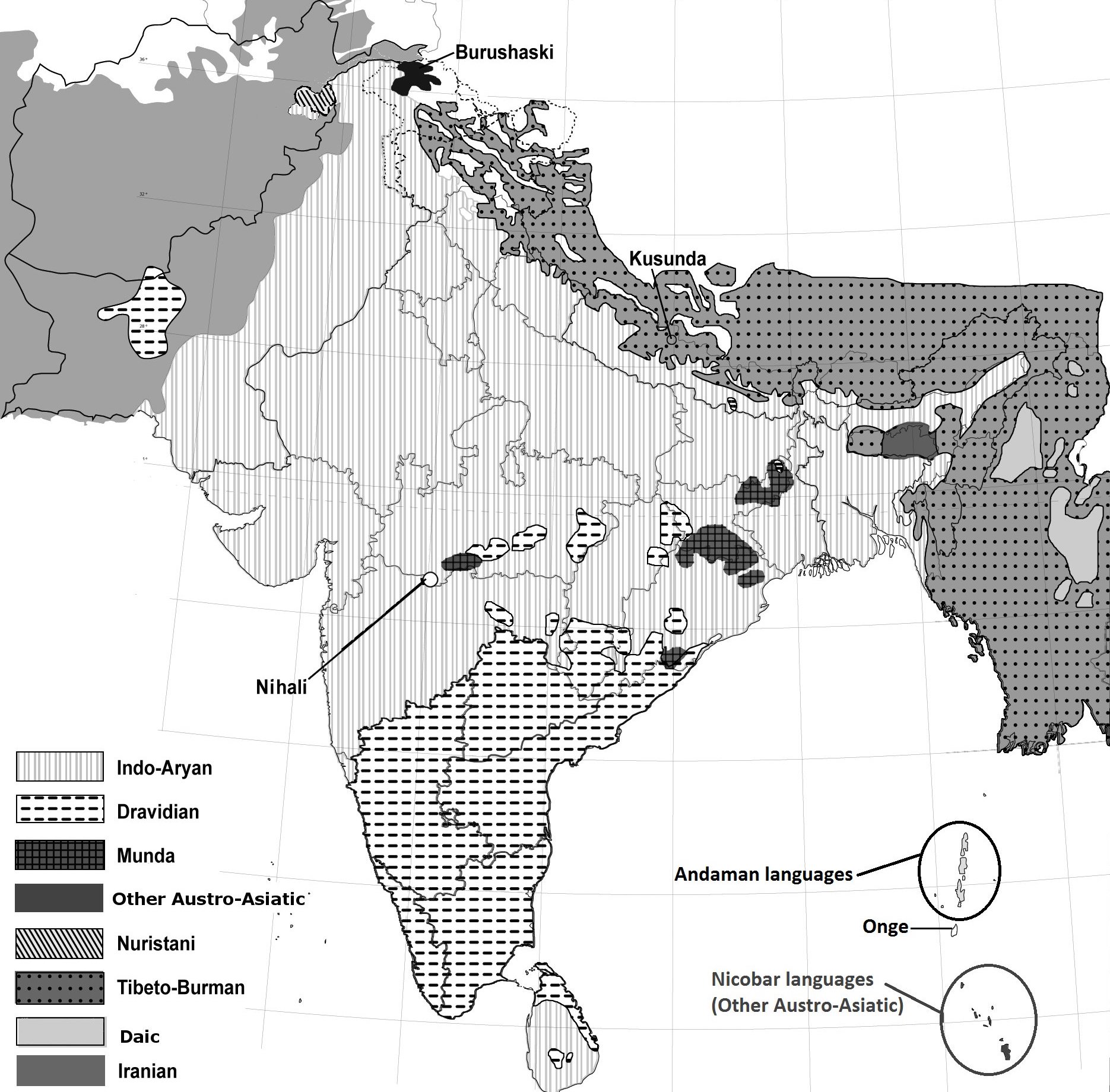}
    \caption{The major language families of South Asia \citep{kolichala}.}
    \label{fig:map}
\end{figure}

We argue that the most basic problem in NLP/CL work on South Asian languages is not data scarcity, but \textbf{data scatteredness}. There is much data to be extracted for even the most endangered languages (e.g., Burushaski, a language isolate of the northwest), from annotated corpora and grammatical descriptions compiled by linguists, \textit{if only} one is willing to wrangle idiosyncratic data formats and digitise existing texts. Thus far, commercial interests and scientific agencies have only intermittently supported the development of language technology for the region---taking a new approach, we propose a research programme from the perspective of computational historical linguistics, outlining current data gathering initiatives in this discipline and potential benefits to other work across NLP.


\section{Background}

\paragraph{Narrowing the low-resource category.} Low-resource languages have recently gained attention in NLP/CL research, both due to the engineering problems of a data-scarce context and also in recognition of the historical focus on English in the field to the detriment of other languages \citep{hedderich-etal-2021-survey,Ranathunga2021NeuralMT}. This has been accompanied by debate on what languages the label encompasses (e.g.~\citealp{hamalainen-2021-endangered}).

In the South Asian context, even Hindi has been labelled low-resource in some recent work. While it is true that for certain tasks a large institutionally-backed language like Hindi can be low-resource, we propose that `low-resource' languages can be better described with two kinds of situations:
\begin{itemize}
    \item \textbf{Data scatteredness}: Data is available (perhaps even abundant), but due to issues in digitisation, cataloguing, and labelling and annotation it has not been leveraged to its full potential.
    \item \textbf{Data scarcity}: Data is not available or very limited to begin with, and without collecting or creating new data we do not have enough to work with.
\end{itemize}

\begin{table}[t]
    \centering
    \small
    \begin{tabular}{p{0.27\linewidth} p{0.6\linewidth}}
    \toprule
    \textbf{Level} & \textbf{Languages} \\
    \midrule
    4:~Underdogs & Hindi \\
    3:~Rising Stars & Urdu, Bengali, Tamil \\
    2:~Hopefuls & Konkani, Marathi, Sanskrit, Punjabi \\
    1:~Scraping-Bys & Malayalam, Bhojpuri, Nepali, Doteli, Gujarati, Newar, Dzongkha, Maithili, Tulu, Kannada, Odia, Kashmiri, Romani, Pashto, Bishnupriya Manipuri, Divehi, Sindhi, Tibetan, Pali, Sinhala, Santali, Assamese, Telugu \\
    0:~Left-Behinds & (several hundred languages) \\
    \bottomrule
    \end{tabular}
    \caption{A brief overview of NLP/CL research progress on South Asian languages grouped by \citet{joshi-etal-2020-state}'s categories.}
    \label{tab:levels}
\end{table}

\paragraph{The state of NLP in South Asia.} So far, initiatives for improving language technology in South Asia have largely focused on data-scattered (not data-scarce) languages with official status and some degree of standardisation. These include cross-lingual projects such as IndicNLPSuite \citep{kakwani-etal-2020-indicnlpsuite}, the EMILLE corpus \citep{mcenery-etal-2000-emille}, and iNLTK \citep{arora-2020-inltk}, and workshops like DravidianLangTech \citep{dravidianlangtech-2021-speech} and WILDRE \citep{wildre-2020-wildre5}. As \cref{tab:levels} shows, only a select few languages benefit from NLP research---even fewer benefit from (commericialised) products like Google Translate or OCR tools. Truly data-scarce langauges (e.g.~Kangri, Tulu) lack instituational status and have been largely unstudied because the challenges are different and harder to surmount.

NLP/CL has proven to be an expansive field as of late. Computational historical linguistics is inextricably linked with computational approaches to fundamental linguistic tasks: corpus building, POS tagging and dependency parsing, morphological analysers, and lexical databases. Work on these has progressed fast for the big languages. For example, Hindi, the highest-resourced South Asian language, has massive hand-annotated dependency treebanks \citep{bhatt-etal-2009-multi}, state-of-the-art neural distributional semantic transformer models \citep{jain2020indictransformers,khanuja2021muril}, and machine translation models to and from English \citep{saini2018neural}.

This is not to say that there are no resources at all for the languages \citet{joshi-etal-2020-state} terms ``the Left-Behinds''. Linguists, for example, have compiled rudimentary treebanks for many languages, simply waiting to be digitised and converted to a multilingual format like Universal Dependencies; these include Palula \citep{liljegren2015palula} and Toda \citep{emeneau1984toda}, which are yet to be the subject of any NLP research work. There are also new treebanks in Universal Dependencies for Kangri, Mandeali, Bhojpuri \citep{ojha-zeman-2020-universal}, and Magahi.

\paragraph{Historical/comparative linguistics.} Historical linguistics is concerned with describing change of all kinds (phonological, morphological, syntactic, etc.) in language over time and the factors (social, cognitive, evolutionary) that contribute to that change. Comparative linguistics aims to use this historical study to relate languages and reconstruct earlier stages and common ancestors of related languages \citep{campbell2013historical}.

The study of historical and comparative linguistics has a long history in South Asia, beginning well before similar threads of inquiry in the Western linguistic tradition, with grammarians like \citet{panini} and \citet{hemacandra} analysing historical and dialectal language from a comparative perspective.

Following the recognition by Western philologists of an Indo-European language family that includes Sanskrit, comparative study of the languages of South Asia began in earnest. As a result, several comprehensive comparative grammars featuring the Dravidian \citep{caldwell,andronov,krishnamurti} and Indo-Aryan families \citep{beames,hoernle,bloch,masica} have appeared in the years since. \citet{emeneau} was the first to posit a South Asian zone of language contact and convergence spanning multiple families. Subsequent work on micro-areal zones has yielded many insights into the nature of linguistic interactions in the region \citep{jharkhand,liljegren,toulmin}.

The sole South Asia-wide linguistic data collection effort ever be undertaken was the \textbf{Linguistic Survey of India}, completed about a century ago \citep{lsi}. To date, there has been no comparable centralised data resource on South Asian languages of its magnitude--covering typological features, the lexicon, and sociolinguistic phenomena.

Data in the earliest comparative works was frequently sourced from high-prestige standard varieties like Delhi Hindi, with progress on studying and collecting data from more localised lects largely proceeding in isolation. Compilation of comparative data continued sporadically throughout the 20th century, resulting in works such as the \textit{Comparative Dictionary of the Indo-Aryan Languages} \citep{CDIAL} and the \textit{Dravidian Etymology Dictionary} \citep{DEDR} which attempt at a more diverse spectrum of language data. Meanwhile, progress on documentation and comparative analysis of the Austroasiatic \citep{munda}, Sino-Tibetan, and isolate languages (e.g. Burushaski, Nihali, Kusunda) of South Asia is still in its infancy. As a consequence, studies drawing upon their data for purposes such as substrate analysis often lack nuance and family-internal consistency.

\section{Ongoing work}
Having established the issue of data scatteredness, the mutual benefit inherent to data collection (for historical/comparative linguistic work and other NLP tasks), as well as possible interesting avenues for future research, we present a compilation of our ongoing projects in this direction, most involving languages that have not been studied in NLP before. 

\begin{figure}[t]
    \centering
    \includegraphics[width=\columnwidth]{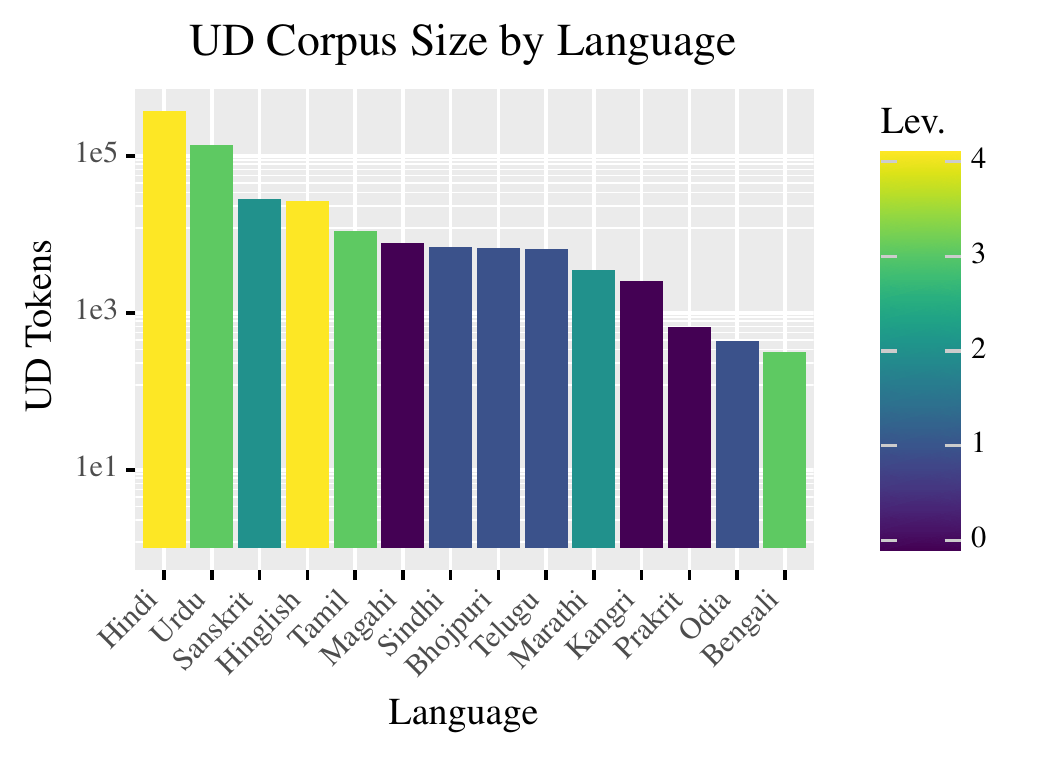}
    \caption{Universal Dependencies corpus sizes, in tokens, for all South Asian languages available thus far. Colors correspond to \citet{joshi-etal-2020-state}'s level categorization.}
    \label{fig:ud}
\end{figure}

\subsection{Dependency treebanks}\label{DT}

\begin{figure*}[t]
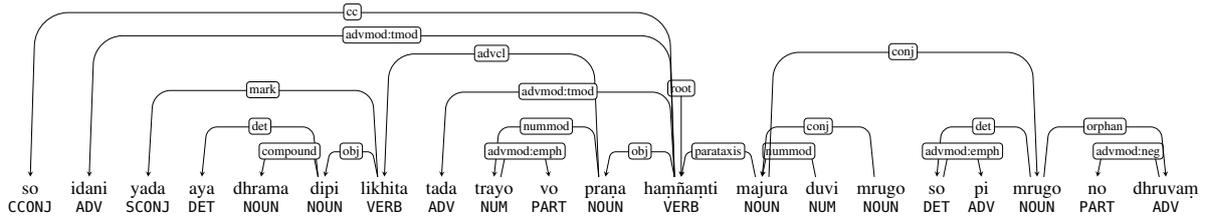

    \centering
    \adjustbox{max width=\textwidth}{
    \begin{dependency}
      \begin{deptext}[column sep=0.2cm]
          so \& idani \& yada \& aya \& dhrama  \& dipi \& likhita \& tada \& trayo \& vo \& praṇa \& haṃñaṃti \& majura \& duvi \& mrugo \& so \& pi \& mrugo \& no \& dhruvaṃ \\
        {\tt CCONJ}\&{\tt ADV}\&{\tt SCONJ}\&{\tt DET}\&{\tt NOUN}\&{\tt NOUN}\&{\tt VERB}\&{\tt ADV}\&{\tt NUM}\&{\tt PART}\&{\tt NOUN}\&{\tt VERB}\&{\tt NOUN}\&{\tt NUM}\&{\tt NOUN}\&{\tt DET}\&{\tt ADV}\&{\tt NOUN}\&{\tt PART}\&{\tt ADV} \\
      \end{deptext}
      \deproot{12}{root}
      \depedge[edge unit distance=1.8ex]{12}{1}{cc}
      \depedge[edge unit distance=1.7ex]{12}{2}{advmod:tmod}
      \depedge[edge unit distance=2.6ex]{7}{3}{mark}
      \depedge{6}{4}{det}
      \depedge{6}{5}{compound}
      \depedge{7}{6}{obj}
      \depedge[edge unit distance=3.7ex]{11}{7}{advcl}
      \depedge[edge unit distance=2.55ex]{12}{8}{advmod:tmod}
      \depedge{11}{9}{nummod}
      \depedge{9}{10}{advmod:emph}
      \depedge{12}{11}{obj}
      \depedge{13}{12}{parataxis}
      \depedge{14}{13}{nummod}
      \depedge{15}{13}{conj}
      \depedge{18}{16}{det}
      \depedge{16}{17}{advmod:emph}
      \depedge{13}{18}{conj}
      \depedge{20}{19}{advmod:neg}
      \depedge{18}{20}{orphan}
    \end{dependency}} \\
    \caption{A sample dependency-parse from the Ashokan Prakrit UD treebank (Shahbazgarhi dialect)}
    \label{fig:depparse}
\end{figure*}
Structured, syntactically-parsed corpora are not only essential for (1) \textbf{downstream NLP tasks} such as information extraction \cite{gamallo-etal-2012-dependency} and semantic role labelling \citep{dep_srl}, but also have the potential to (2) aid quantitative \textbf{comparative and historical linguistic study}. Parsing according to several formalisms is possible, though dependency formalisms in particular are better equipped to handle the flexible word-order characteristic of many South Asian languages (assuming the parsing algorithm used adequately handles non-projective dependency trees\footnote{
For vertex set $V$, weighted edge set $E \subseteq \{i \xrightarrow{w} j \ \vert \ i, j \in \mathbb{R}, w \in \mathbb{R} \}$, and root $\rho \in V$, let $G = (\rho, V, E)$ be a rooted weighted directed graph. A \textbf{dependency tree} is a spanning subgraph $D = (\rho, V, E'), \ E' \subseteq E$   subject to the following well-formedness constraints \cite{zmigrod-etal-2020-please}:
\
\begin{itemize}
    \item [(C1)] Each non-root vertex of $D$ has one incoming edge
    \item [(C2)] $D$ is acyclic 
    \item [(C3)] Root $\rho$ of $D$ has exactly one outgoing edge 
\end{itemize}

In other words, dependency trees are \textit{arborescences} (directed, rooted trees) equipped with the root constraint (C3). Graph-based parsing algorithms find the optimal dependency tree $D^{*}$, that is, the dependency tree $D$ with maximum total edge weight in the set of all possible dependency trees $D(G)$, for a given sentence  (maximum weight spanning arborescence). A \textbf{treebank} is a corpus of such dependency trees.
}) \cite{palmer2009hindi}.

Multilingual dependency formalisms such as Universal Dependencies (UD) \cite{nivre-etal-2016-universal} have established consistent guidelines for the annotation of binary dependency relations, morphology, and other linguistic features, resulting in the recent appearance of treebanks for several data-scarce languages of the region (Bhojpuri, Kangri, etc.) as well as their older diachronic stages (Vedic and Classical Sanskrit).

Towards the second goal listed above, \citet{farris-arora-2022} compiled a UD treebank for the Ashokan Prakrit dialect continuum--a parallel corpus of 14 pillar/rock inscriptions in six Middle Indo-Aryan (MIA) dialects dating back to the 3rd c.~BCE. As the first study of MIA from a computational perspective, this work calls for an analysis of Indo-Aryan regional fragmentation through dialectometry, approaching contentious linguistic issues with statistical arguments curated using treebank data.

In a similar vein, we are currently working towards filling other chronological gaps in corpora (e.g. the Old Sinhala Sīgiri Graffiti of the Early New Indo-Aryan stage) through treebanking \textit{in parallel} with their modern stages (e.g. Sinhala). To the best of our knowledge, we are unaware of any studies involving such \textbf{diachronic transfer} frameworks, where knowledge transfer between two historically-separated stages of the same language can be used to dependency-parse a given stage using resources from the other.  Other historically-attested langauges we plan to include in this pipeline include Old Kashmiri, Old Maldivian, and Old Tamil.

In terms of modern South Asian languages, there has been recent diversification from combined efforts, such as an upcoming dependency parsing shared task at the WILDRE 2022 workshop based on new treebanks \citep{nallani-etal-2020-fully,ojha-zeman-2020-universal}.

\paragraph{Multilingual dependency parsing.} More broadly, we are interested in cross-lingual transfer models \cite{duong-etal-2015-cross, guo-etal-2015-cross, schuster-etal-2019-cross} as a means of expediting dependency parsing for data-scarce South-Asian languages. A similar approach for Uralic languages is \cite{lim-etal-2018-multilingual}. They propose a dependency-parsing model for North Saami and Komi using annotated corpora and bilingual word-embeddings from high-resourced genetically related (Finnish) and typologically similar (Russian) languages, without the requirement of extensive parallel texts for training. They conclude that while genetically related pairs (Komi--Finnish, North Saami--Finnish) allow for highly efficient parsing, pairs of unrelated languages in contact (Komi--Russian) also provide valuable input for further correction. Given the languages of South Asia exhibit common typological features by virtue of sharing a linguistic area, treebanking efforts will undoubtedly beneft from a multilingual dependency parsing approach. Languages like Sindhi, Punjabi, and Sinhala, which have genetic relatives and contact languages that are comparatively more resourced, are our immediate targets for such efforts.



\subsection{\textit{Jambu} etymological database}
\begin{figure}[t]
    \centering
    \includegraphics[width=\columnwidth]{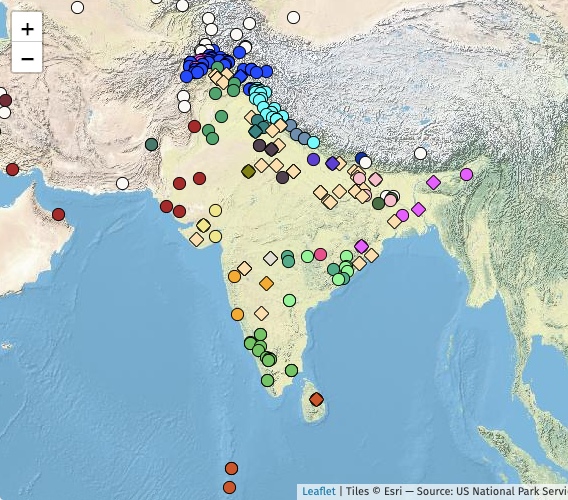}
    \caption{A map of languages included in Jambu, colour-coded by subfamily designation with point-geometry variation by diachronic stage.}
    \label{fig:jambumap}
\end{figure}
\begin{figure*}[t]
    \centering
    \includegraphics[width=\textwidth]{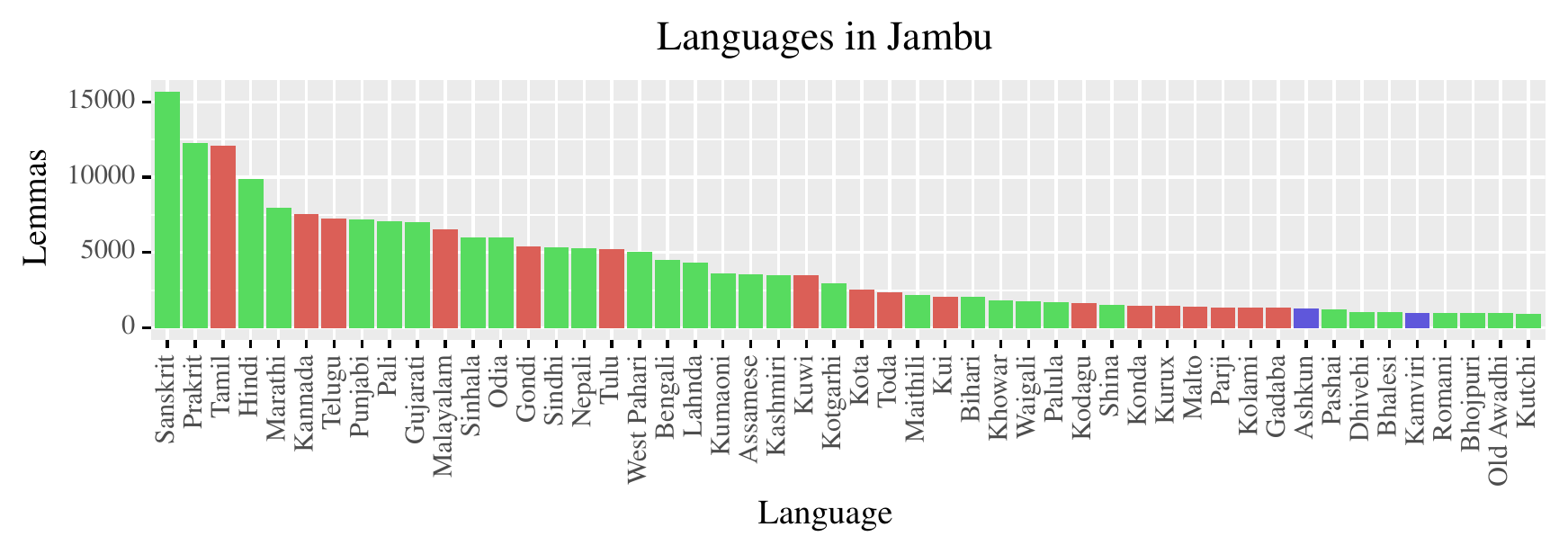}
    \caption{Top 50 languages by number of lemmas included in the Jambu database, colour-coded by language family (green = Indo-Aryan, red = Dravidian, blue = Nuristani).}
    \label{fig:jambu}
\end{figure*}
One of our major efforts in data-collection for the region has been the \textbf{Jambu} project. Jambu is a compiled cognate lexicon of all South Asian languages, cutting across phylogenetic groupings and historical language stages. It has a web interface online at \url{https://neojambu.glitch.me/}. It includes data parsed and compiled from the University of Chicago's Digital Dictionaries of South Asia project \citep{CDIAL,DEDR}, existing web databases \citep{liljegren,strand}, and individual articles and theses \citep{toulmin,kvar}, totalling 294 lects and 202,653 lemmas. Some of these sources have been used in previous work on South Asian historical linguistics, e.g.~\citet{cathcart-rama-2020-disentangling,cathcart-2019-toward,cathcart-2019-gaussian,cathcart-probabilistic}---this is the first attempt to consolidate them. Note some previous work in this direction: while the SARVA project \citep{sarva} did not reach fruition, a searchable database of Dravidian cognates was developed by Suresh Kolichala under its auspices.\footnote{\url{http://kolichala.com/DEDR/}} 




Past etymological research in South Asian languages was primarily focused on internal comparisons within linguistic families. Unknown etyma was often blindly attributed to Dravidian or Munda without comprehensive cross-linguistic analyses.\footnote{Recent comparative work on Munda and Indo-Aryan contact such as \citet{ivani} in general find very limited influence of Munda, restricted primarily to the (eastern) Indo-Aryan languages in close proximity with them. Prior work had a tendency to exaggerate the impact of Munda to explain unusual features of other Indo-Aryan languages; notably, \citet{witzel1999substrate}, who advocated for a historical `Para-Munda' family that influenced Indo-Aryan as far as in the northwest, the historical location of Rigvedic Sanskrit.} In fact, we find a large number of common words in languages of several families with uncertain origin, possibly substrate loans from undocumented languages.\footnote{
Dr. Felix Rau (p.c.) terms these unattested substrate(s) `the \textbf{big X} of South Asian linguistic history', and other (possible same) substrate(s) responsible for words reconstructable to Proto-Munda without secure cognates in other Austroasiatic branches `the \textbf{big Y}'.
} 
In order to provide reliable data for the robust reconstruction of the history of the ancient linguistic contact, a comprehensive South Asia-wide linguistic data is desideratum.

\paragraph{Consolidating Indo-Aryan data.} While \citet{CDIAL} and its supplements remain the undisputed gold standard for Indo-Aryan comparative etymologies, many later works on individual languages have considerably expanded our knowledge of cognate relations in underdocumented languages; e.g.~\citet{liljegren,toulmin,zoller2005grammar}. Inclusion of data from these newer works is ongoing. We also expanded coverage of the isolated and linguistically archaic Nuristani lects \citep{strand}, which are contended not to be Indo-Aryan---comparative lexical data will help cement their exact phylogenetic status.

\paragraph{Updates to Dravidian data.} A \textit{Dravidian Etymological Dictionary} published by \citet{DEDR} (2nd edition; abbreviated DEDR) remains the latest effort to gather etymological data on Dravidian. Although \citet{krishnamurti} provides reconstructions for about 500 entries, systematic historical reconstruction for all known cognates of Dravidian is still pending. \citet{supplement_dedr} published an update to the DEDR utilizing new data on several non-literary languages that became available after 1984.

Recent fieldwork on several non-literary languages have produced grammars with new vocabulary lists, providing rich data to be updated in DEDR. In addition, several dictionaries with attempted etymologies for many literary languages have appeared since 1984, and can become a source for the realignment of cognates as well as new additions. 

\paragraph{Cognate databases in NLP.} The obvious benefit of cognate databases for upstream NLP tasks is for data-scarce languages that lack adequate corpora on the web. Similar work in this area is the pan-lingual CogNet \citep{batsuren-etal-2019-cognet}, and also earlier WordNets \citep{miller1995wordnet}. Cognate data can be used for transfer learning, where a data-scarce language can map onto existing models for higher-resource languages, such as a distributional semantic model which generally requires massive corpora to train \citep{sharoff-2017-toward}. Typological data in general offers modest improvements in performance on a variety of NLP tasks \citep{ponti-etal-2019-modeling}.

\paragraph{Unified transcription.} Since many languages of South Asia are unwritten or are lacking standardised orthographies (even in their respective linguistics works), we developed a preliminary system for phonemic transcription of all South Asian languages, which all our cognate data will be converted to.  For cognate identification and reconstruction work (both by humans and using NLP tools), a unified phonemic representation is important. This system combines features of the International Alphabet of Sanskrit Transliteration (IAST)\footnote{\url{https://en.wikipedia.org/wiki/International_Alphabet_of_Sanskrit_Transliteration}} with IPA and Americanist phonetic transcription systems. Future work will outline it in depth, along with examples of its focus on cross-family diacritical consistency.

\subsection{Historical linguistic analyses}
One of our main objectives for building extensive comparative lexical and grammatical databases is to ensure credible data from up-to-date, modern sources are available to researchers working on comparative and diachronic linguistics in the South Asian linguistic area. Historical linguistics work needs data, and in South Asia too much work has progressed without including data from non-standardised (even if documented) languages, to the detriment of our understanding of South Asian linguistic history post-Sanskrit \citep{kurdish}.

Below, we highlight two such projects we are currently engaged in involving three data-scarce languages of northern Pakistan: Burushaski, Gawri, and Torwali \citep{torwali2018}.

\subsubsection{Gawri tonogenesis and UniMorph}
The languages of northern Pakistan have been synchronically analyzed to have phonemic tonal contrasts. \citet{baart2003} has classified such tonal languages into three broad groups based on the type of tonal contrast displayed:
\begin{itemize}
\item \emph{Shina-type}: Shina varieties, Palula, Indus Kohistani (all Indo-Aryan), Burushaski (isolate) etc.
\item \emph{Punjabi-type}: Punjabi, Hindko, some Gujari varieties, extending into the Himachali languages of northern India, as well as Kishtwari\footnote{Not mentioned in \citet{baart2003}, but independently identified by one of the present authors.}, which is usually classified as a divergent dialect of Kashmiri (all Indo-Aryan).
\item \emph{Kalami-type}: Gawri (Kalami), Kalkoti and Torwali (all Indo-Aryan) and possibly other undiscovered varieties of the area.
\end{itemize}
To these, one may also add the simpler accentual systems of Kalasha-mon \cite{petersen2015} and Khowar \citep{liljegren2017}, which we term \emph{Chitrali-type}.

The tonal system and the historical mechanism of tonogenesis is broadly understood for Punjabi proper and some Hindko varieties \citep{shackle1980, bashir2019, bhatia2013}, but specifics for individual varieties further east (Kishtwari and Himachali) remain underdescribed \citep{hendriksen1986, kvar}. This system arises primarily from the disappearance of phonemic breathy voice, but the phonetic specifics differ from language to language. The Shina-type tonal system is both the best described and the best understood diachronically. It continues the Vedic (hence Indo-European) pitch-accent system subject to later changes necessitated by regular apocope \citep{liljegren2008, liljegren2016, kuemmel2015}. Vedic pitch-accent is also partly continued by the Chitrali-type accentual system \citep{petersen2012},  though less conservatively.

The tonal diachrony of the Kalami-type system, on the other hand, has not yet been fully understood. Part of the reason is that this system is considerably more complex than the other three accentual systems, contrasting as many as five distinct tonemes \citep{baart1997, lunsford2001, liljegren2013}. In ongoing work, based on the Gawri data compiled from \citet{baart1997, baart1999, sagar2004, baart2004}, we are investigating the origin of the system, and will be appended in the future by Torwali data we are now collecting.

\paragraph{Morphology in NLP.} 
\begin{figure}[t]
    \centering
    \includegraphics[width=\linewidth]{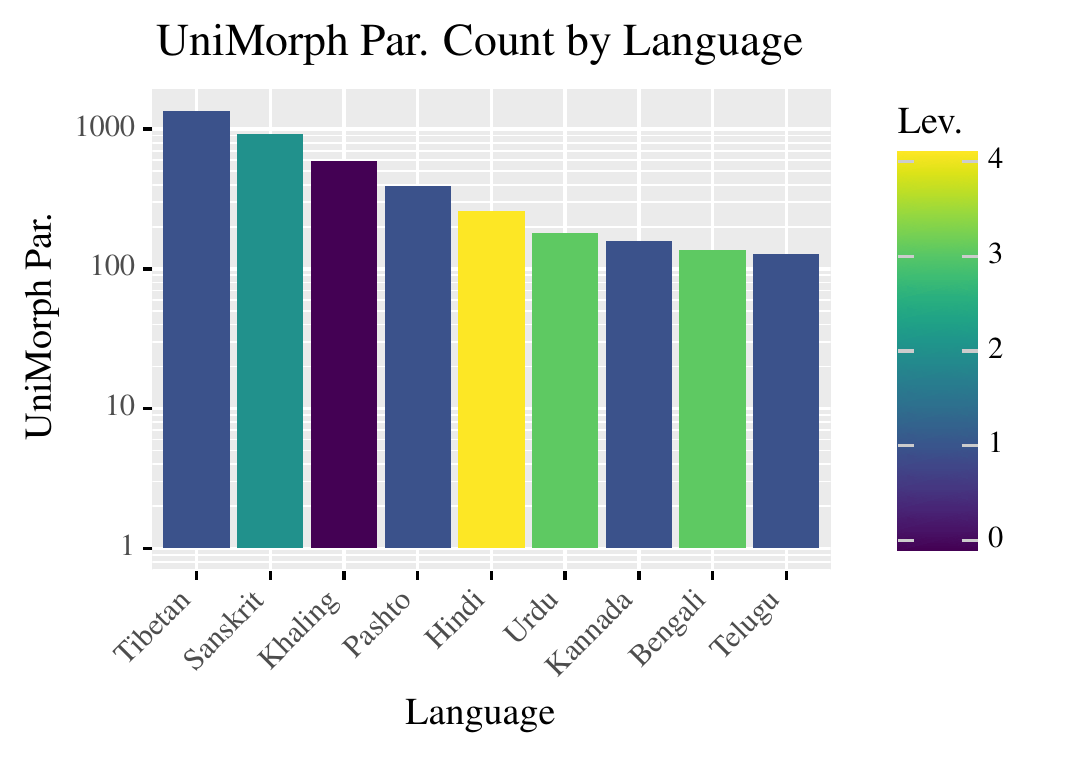}
    \caption{UniMorph paradigm counts for all South Asian languages available thus far. Colors correspond to \citet{joshi-etal-2020-state}'s level categorization.}
    \label{fig:unimorph}
\end{figure}
In addition to working out the history of the Kalami-type tonal system, we intend to incoporate our annotated lexical dataset into the \textbf{UniMorph} database \citep{kirov-etal-2018-unimorph}. The morphology of Gawri and Torwali marks gender, number and case for nouns and adjectives primarily by tonal changes and vowel alterations (historical umlaut) unlike other Indo-Aryan languages which use suffixation, though they still encode much the same categories and do not behave any different syntactically either. This makes them prime targets for testing out computational methods for morphological analysis, especially to compare performance vis-\`{a}-vis a related language like Hindi that has a similar grammar but different morphological profile.

UniMorph has only a few South Asian languages thus far, as shown in \cref{fig:unimorph}---this is part of a broader project to expand coverage in the region, using existing morphological data stored in analysers (e.g.~for Sindhi, \citealp{motlani-etal-2016-finite}) and grammars (e.g.~for Palula, \citealp{liljegren}). In this vein, we also mention that UniMorph has only a handful of languages that signal morphological alterations tonally. So, our contribution will also improve typological diversity in the database to a considerable extent.

\subsubsection{Proto-Burushaski reconstruction}\label{PBR}
Our understanding of the linguistic pre-history of South Asia is heavily reliant on disciplined studies of the histories of the non-Indo-European languages of the subcontinent. This is primarily because while we do have reliable estimates on the time-frame of Indo-European migration into the subcontinent, for the families endemic to the region (including isolates) analogous dating is not possible.

Burushaski, spoken in a few mountain valleys of the Karakoram, is among these endemic languages of South Asia. It has attracted quite a bit of scholarly attention since its academic discovery as it stands out both typologically and genealogically in its current neighborhood (cf. the latest descriptive grammars \citet{berger1974, berger1998, munshi2019, yoshioka2012}). The history of the language and its speakers is virtually unknown until the first linguistic documentation in the mid-nineteenth century. The first secure pre-modern attestation of Burushaski speakers is in Tibetan chronicles dating from the ninth century where a people \textit{bru-\`{z}a} or \textit{bru-\`{s}a} to the west of Tibet find mention \citep{jaeschke1881}.\footnote{We are grateful to Dr. Diego Loukota (p.c.) for informing us that a short text in the \textit{bru-\`{s}a} language is also attested in Tibetan records with translation in Sanskrit. We are, however, not aware of any scholarly attempt to interpret said text through modern Burushaski.}\footnote{It is also possible that an older ethnonym recorded as Sanskrit \textit{m\={u}ja-}, \textit{maujavata-} and Avestan \textit{mu\v{z}a-} refer to the same people but that is harder to establish.}

As of now, both major varieties of Burushaski are well-documented, but there has been precious little comparative work done. The dictionaries in \citet{berger1974, berger1998} lay the  foundation of comparative studies by identifying several layers of potential loans in the language, cf. also \citet{rybatzki2010}. Conversely, potential Burushaski interaction with and influence on the older stages of Indo-Iranian have been explored in \citet{tikkanen1988, kuemmel2018}, the former mainly dealing with how Burushaski broadly fits into the South Asian linguistic zone. A handful of Burushaski loans in Purik Tibetan are identified in \citet{zemp2018}, not all of them convincing, and \citet{stebline1999} contains shared lexemes with Wakhi. More speculative are the claimed Burushaski loans in (Proto-)Romani collected in \citet{berger1959}, believed to be borrowed before the Roma migrated westward toward Europe (presumably) through Burushaski territory.

However, all these studies share a common drawback in that we do not yet have a principled way of identifying Burushaski lexemes or grammatical features. A first step toward this goal is \citet{Holst2014}, where the author attempts an internal reconstruction of Burushaski through a comparative lexical and morphological study of the two main dialect groups of Yasin and Hunza--Nager. Holst's work, though, is still just a preliminary investigation and there is much to be added and improved on. In particular, the book does not undertake a systematic study of loanwords to and from neighboring languages as previous areal studies involving Burushaski have, nor does it exhaustively utilize the descriptive literature available resulting in a few avoidable but significant errors of interpretation \citep{Munshi2015}. This is a major shortcoming because external comparisons are a vital component to reconstructing the histories of language isolates and smaller families, cf. \citet{trask2013} for Basque and \citet{nikolaeva2011} for Yukaghir, among others.

\paragraph{Computational reconstruction.} We have already started a principled reconstruction of Proto-Burushaski building on Holst's work, but utilizing more sources and laying a greater emphasis on loanword etymologizing and chronologizing. Our databases, compiled from available lexical and descriptive sources, are intended to aid this goal of comparative analysis, as well as to make data from Burushaski and neighboring languages available to other researchers.

Proto-language reconstruction is an interesting task in computational historical linguistics, and so far work has been under way in a supervised setting on known, high-quality cognate data across related languages, e.g.~on Romance languages \citep{ciobanu-dinu-2018-ab,meloni-etal-2021-ab}.

\paragraph{Low-resource dependency corpora.} In addition, starting with annotated texts from descriptive grammars, we plan to build a dependency treebank for Burushaski as described in \cref{DT}. Burushaski is a low-resourced language in the sense that its domain of use is very restricted and there is no readily available internet corpus one can subject to sophisticated (computational) linguistic analyses automatically.

However, as mentioned before, there has been a steady stream of quality descriptive work on it and all published grammars come with a wealth of oral texts one can build a functional corpus with---indicating some data-scatteredness that can be leveraged.

\section{Future Work}
The data resources we are in the process of compiling for South Asian languages will enable a variety of research to be conducted into language history. We lay out some of the immediate potential pathways for this further research in hopes of stimulating work in this area.

\subsection{Substrate studies and language history}
A substrate language is one that loans words into a language of higher prestige. A perennial question in South Asian language history for at least a century has been the Indus Valley Civilisation inscriptionary corpus, and the problem of deciphering it (if it even encodes a language) and whether it belongs to a known language family of South Asia or something else entirely \citep{farmer2004collapse,fairservis1983script}. Notably, in the mid-20th century a team of Finnish and Soviet linguists and computer scientists claimed evidence that the Indus inscriptions represent a Dravidian language \citep{parpola1986indus}.

Recent computational information-theoretic work also suggests language-like properties in the text, a subject of subsequent vociferous debate \citep{rao2009entropic,rao-etal-2010-commentary}. A serious issue is that we do not have sufficiently diverse data from modern languages of the region against which to compare any purported decipherments of the Indus script (e.g.~Proto-Dravidian reconstruction is as of now still in a preliminary stage), and thus even if the Indus language provided any substrate loans into modern families, we would be unable to comprehensively list out possible candidates. The \textit{Jambu} database can help inform research on substrate contact in the languages of the region.

\subsection{Text digitisation and OCR}
One of the major bottlenecks in compiling existing linguistic data on South Asian languages is that it remains machine-unreadable. For example, many linguistics theses completed at Indian universities have recently been digitised and uploaded to Shodhganga\footnote{\url{https://shodhganga.inflibnet.ac.in/}}, but most are scanned images in PDF format. Optical character recognition (OCR) of such texts also requires difficult parsing of diacritics and low-resource scripts.

A recent initiative to digitise old linguistic data is the digitisation of the Linguistic Survey of India \citep{lsi} under the project \textit{South Asia as a linguistic area? Exploring big-data methods in areal and genetic linguistics} \citep{borin,borin2,borin-etal-2014-linguistic}. Using OCR and subsequent information extraction from the text, \citeauthor{borin} have shown that ``old'' data still has much to tell for the computational study of typology and comparative linguistics.

Future work on extracting data from non-digitised South Asian language sources will have to use OCR, possibly a neural model finetuned for the purposes of our domain on a platform like Transkribus \citep{kahle2017transkribus}.
\subsection{Fieldwork initiatives}
\citet{hamalainen-2021-endangered}, calling for the NLP community to make a consistent distinction between ``endangered'' and ``low-resource'' languages, implores researchers to `stop complaining about how low-resourced [a language] is, [and] get up and gather the data.'

In response to this call, we announce several currently-underway (online) fieldwork/data elicitation efforts for Indo-Aryan languages that are both endangered and data-scarce. These include Kholosi, Poguli, Kishtwari, Bhaderwahi, Torwali, and certain divergent dialects of Maldivian (e.g. Huvadhoo). By virtue of their geographical spread (Northern India/Pakistan, Iran, Maldives), linguistic data collected from these languages will further enable the consturction of typologically viable datasets for both NLP and computational historical linguistic tasks.

\section{Conclusion}
In this paper, we gave an overview of the state of NLP in South Asia with a special focus on historical--comparative linguistics, a research programme of which we believe will help address the issue of data scatteredness. South Asian languages are not obliged to remain low-resource (in the NLP sense), and have plenty of speakers who would like access to and would benefit from language technologies, along with a multitude of raw linguistic resources that can be used to cultivate them. Incentives have not been in place to support those demands, however, so we suggest an alternative route founded in linguistic research to gather data.

Collective efforts have had great success recently in NLP---besides institutional efforts like the Stanford Center for Research on Foundation Models \citep{bommasani2021opportunities} and HuggingFace's BigScience Workshop\footnote{\url{https://bigscience.huggingface.co/}}, there are grassroots organisations like MaskhaneNLP for African languages \citep{nekoto-etal-2020-participatory} and AI4Bharat \citep{kakwani-etal-2020-indicnlpsuite} that are working towards improving resource availability. Our proposals in this paper are the first seeds of a programme similar in spirit, motivated by a dual interest in understanding South Asian language history and remedying inequalities in technological availability.

\bibliography{anthology,custom}
\bibliographystyle{acl_natbib}



\end{document}